\documentclass[letterpaper]{article} 
\usepackage{aaai24}  
\usepackage{times}  
\usepackage{helvet}  
\usepackage{courier}  
\usepackage[hyphens]{url}  
\usepackage{graphicx} 
\usepackage{natbib}  
\usepackage{caption} 
\usepackage{algorithm}
\usepackage{mdframed}
\usepackage{amsmath}
\newmdenv[backgroundcolor=gray!20]{myframe}

\usepackage{fancyvrb,newverbs,xcolor,subcaption}

\definecolor{cverbbg}{gray}{0.94}

\usepackage{algorithm}
\usepackage{algpseudocode}
\usepackage{newfloat}

\usepackage{listings}
\DeclareCaptionStyle{ruled}{labelfont=normalfont,labelsep=colon,strut=off} 
\floatstyle{ruled}
\newfloat{listing}{tb}{lst}{}
\floatname{listing}{Listing}
\usepackage{booktabs}
\nocopyright

\title{DemOpts: Fairness corrections in COVID-19 case prediction models}
\author{
    Naman Awasthi\\
    Saad Mohammad Abrar,\\
    Daniel Smolyak,\\
Vanessa Frias-Martinez
}
\affiliations{
    University of Maryland, College Park\\
}

\usepackage{bibentry}

\begin{document}

\maketitle

\begin{abstract}
COVID-19 forecasting models have been used to inform decision making around resource allocation and intervention decisions e.g., hospital beds or stay-at-home orders. State of the art deep learning models often use multimodal data such as mobility or socio-demographic data to enhance COVID-19 case prediction models. 
Nevertheless, related work has revealed under-reporting bias in COVID-19 cases as well as sampling bias in mobility data for certain minority racial and ethnic groups, which could in turn affect the fairness of the COVID-19 predictions along race labels. 
In this paper, we show that state of the art deep learning models output mean prediction errors that are significantly different across racial and ethnic groups; and which could, in turn, support unfair policy decisions. 

We also propose a novel de-biasing method, DemOpts, to increase the fairness of deep learning based forecasting models trained on potentially biased datasets. Our results show that DemOpts can achieve better error parity that other state of the art de-biasing approaches, 
thus effectively reducing the differences in the mean error distributions across more racial and ethnic groups. 

\end{abstract}

\section{Introduction}
Forecasting the number of COVID-19 cases, hospitalizations or deaths is crucial to inform decision making. For example, COVID-19 forecasts can be used by hospitals to evaluate medical needs and required resources such as supplies or beds; or by public health officials to inform closure policies at various geographical scales. In the US, COVID-19 forecasts have been used at the state and county levels to inform social distancing or masking, such as the publicly available forecasts on the COVID-19 Forecast Hub that the CDC routinely uses in their communications \cite{CDC2023, forecasthub}.

Related work for the past three years has shown a diverse variety of COVID-19 forecasting approaches, from compartmental models \cite{compartmental1,compartmental2} to statistical \cite{stats1,stats2} or deep learning methods \cite{deep1,deep2,deep3,deep4}. These models, always trained with past COVID-19 cases publicly available
from, for example,  NYT or JHU \cite{nytimes2021covid,JHU}, frequently use complementary datasets with the objective of improving forecasting accuracy. In fact, analysing the forecasting models from over $50$ teams in the COVID-19 Forecast Hub, $39\%$ use demographic data - either directly from the ACS or indirectly via community vulnerability indices like the CCVI \cite{ccvi}; and $52\%$ of the models incorporate human mobility data from Safegraph, Google or Apple, among others \cite{google2022covid,apple2022covid,descarteslabs2023covid}. 

The majority of publications focused on COVID-19 case prediction have reported results around the accuracy of the models \textit{i.e.,} minimizing the difference between the predicted cases and the actual number of cases reported. Nevertheless, prior work has shown that the accuracy of COVID-19 predictions can depend on various social determinants, including race or ethnicity \cite{gursoy2022error}, income, or age \cite{Erfani_Frias-Martinez_2023}, revealing worse performance for protected attributes and pointing to a lack on COVID-19 predictive fairness that can affect resource allocation and decision making. 
This lack of predictive fairness might be related to bias in the datasets used to train the model \textit{i.e.,} 
bias in COVID-19 case reporting or bias in mobility data. 
In fact, prior work has shown COVID-19 case bias due to under-reporting issues in minority communities whereby missing racial data or misclassified race has been a source of errors \cite{douglas2021variation} as well as inadequate testing for minority groups across the US, such as Hispanic/Latino communities \cite{ama}.
Additionally, prior work has also revealed sampling bias in mobility data with Black and elder communities being under-represented because of the way mobility data is collected (via smart phones and mobile app use) \cite{coston_bias_mob}.

Given the presence of bias in the training datasets frequently used by COVID-19 forecast models, and prior work demonstrating that COVID-19 prediction accuracy can vary across social determinants, we posit that it becomes critical to devise methods to prevent data biases from percolating into the COVID-19 forecasts so as to guarantee fair decision making based on case predictions.  
Mitigating bias in COVID-19 forecast models can be done through pre-processing or in-processing approaches \textit{i.e.,} via bias mitigation in the training datasets, applying correction methods to COVID-19 counts \cite{angulo2021estimation, jagodnik2020correcting}; or via de-biasing methods embedded in the predictive models that attempt to reduce data and model bias during training \cite{Yan_Seto_Apostoloff_inproc,inproc_Socmedia,covidclinicfair1,covidclinicfair2}. 
In this paper, we focus on in-processing approaches given their scarcity in the COVID-19 literature,
and propose \textit{DemOpts} (Demographic Optimization) a de-biasing method designed to achieve COVID-19 case prediction error parity across racial and ethnic groups in the context of deep learning models \textit{i.e.,} guarantee that county prediction errors are not significantly different across racial and ethnic groups. Although there exist a diverse set of COVID-19 predictive approaches, we focus on deep learning models, because these are the most frequently used models in the machine learning community \cite{meraihi2022machine}; and narrow down our choice to transformer-based architectures in particular, because they are state of the art in time series predictions \cite{lim2021temporal}.  

The main objective of DemOpts is to improve the fairness of the COVID-19 case predictions at the county level by achieving error parity in a regression setting \cite{gursoy2022error}. 
DemOpts proposes a novel de-biasing approach that leverages county racial and ethnic data during training to modify conventional deep learning loss functions so as to penalize the model for statistically significant associations between the predictive error and the race or ethnicity distribution of a county. 

Like state of the art de-biasing methods for regression settings (such as Individual~\cite{berk2017convex},
Group~\cite{berk2017convex} and Sufficiency-based fairness correction~\cite{sufficiencyabhin}) 
DemOpts can work in multimodal contexts, allowing for deep learning models to be trained with different types of input data besides the COVID-19 cases, including the use of mobility or demographic data, which are frequently used in COVID-19 prediction models. 
However, unlike state of the art de-biasing methods for regression,
DemOpts is designed to de-bias predictions based on the relationship between the prediction errors and the percentage of racial and ethnic groups in that county, effectively considering multiple protected groups per county in the de-biasing process, instead of assigning a county to a unique protected race or ethnicity.

Thus, the main contributions of this paper are:
\begin{itemize}
    \item We present DemOpts, a novel de-biasing method for deep learning architectures, that attempts to increase the fairness of the COVID-19 county case predictions by achieving error parity \textit{i.e.,} guarantee that prediction errors are similar across racial and ethnic groups. 
    
    \item The DemOpts architecture is designed 
    to optimize for error parity across race and ethnicity using a novel multi-label approach that allows each county to be characterized by its own racial and ethnic group distribution during the de-biasing process, instead of by a unique label. 
    
    \item We propose a novel evaluation protocol for the COVID-19 context and we show that:
    (i) state of the art COVID-19 county case prediction models based on transformer architectures lack error parity when no de-biasing method is applied \textit{i.e.,} prediction errors are statistically significantly different across racial and ethnic groups; (ii) DemOpts applied to transformer-based architectures improves the error parity of COVID-19 county case prediction models, increasing the similarity between mean prediction errors across racial and ethnic groups, and 
    (iii) DemOpts de-biasing approach performs better than existing de-biasing methods for regression settings, namely, individual fairness correction, group fairness correction and sufficiency-based penalty for fairness correction.
    
    \end{itemize}

The rest of the paper is structured as follows. We first discuss the related literature, followed by a description of the DemOpts method to achieve error parity in transformer based architectures, and its evaluation protocol for the COVID-19 context, including two metrics to measure error parity. We finalize the paper presenting the evaluation of DemOpts: first describing the datasets used, and then showing how DemOpts improves fairness prediction by increasing error parity across racial and ethnic groups.

\section{Related Literature}

In this section, we cover three areas that are of relevance to the research proposed in this paper: deep learning models to forecast time series data, the presence of bias in COVID-19 datasets used by COVID-19 case forecasting methods; and approaches to measure and improve the fairness of predictions in regression settings.

\subsection{Deep learning based Forecasting models}
Deep learning models have started to become popular in time series prediction tasks. The available methods include, (i) Autoregressive models, which are modifications of recurrent neural networks (RNNs) such as Long Short Term Memory (LSTM) Networks or Gated Recurrent Networks (GRN)\cite{LSTMSchmidhuber}; (ii) Graph-based neural networks which encapsulate spatio-temporal aspects of data in implementations such as Graph Attention Networks (GANNs)\cite{GANN}, Spatio-temporal Graph Convolution network (ST-GCN)\cite{STGCN}, NBConv \cite{nbconv} or GGConv\cite{ggconv}; and (iii) transformers, which have gained success in various applications such as computer vision \cite{vit,sam}, natural language processing \cite{GPT1,BERT}, speech \cite{speechtransformer} or tabular data \cite{tapex,TUTA,TABERT}. There is a large body of work that utilizes Transformer-based architecture models \cite{transformer} to forecast time series with state of the art performance including LogTrans (Li et al., 2019)\cite{LogTrans}, Informer (Zhou et al., 2021) \cite{informer}, Autoformer (Wu et al., 2021) \cite{autoformer}, FEDformer (Zhou et al., 2022)\cite{fedformer}, Pyraformer (Liu et al., 2022)\cite{pyraformer}, and PatchTST\cite{PatchTST}. In this paper, we specifically focus on the Temporal Fusion Transformer architecture (TFT) \cite{TFT} since it allows us to easily incorporate exogenous variables (like mobility data) as well as static variables (like demographic data) on top of the COVID-19 time series.

\subsection{Bias in Mobility and COVID-19 Data}
Human mobility data has been used to characterize human behaviors in the built environment 
~\cite{vieira2010querying,hernandez2017estimating,frias2013cell,rubio2010human,wuspatial}, 
for public safety~\cite{wu2022enhancing,wu2023auditing}, during epidemics and 
disasters~\cite{wesolowski2012quantifying,bengtsson2015using,hong2017understanding,isaacman2018modeling,ghurye2016framework,hong2020modeling}, as well as to support decision making for socio-economic development
~\cite{frias2010socio,fu2018identifying,frias2012mobilizing, hong2016topic,frias2012computing}.
During the COVID-19 pandemic, human mobility has also played a central role in driving decision making, acknowledging the impact of human movement on virus propagation \cite{arik2020interpretable, lucas2023spatiotemporal, le2020neural, Erfani_Frias-Martinez_2023, badr2021limitations, abrar2023analysis}.

Bias in mobility and COVID-19 data is being increasingly discussed due to the exponential growth of COVID-19 forecasting models in the literature, and to publicly available mobility data. 
Mobility data has been reported to suffer from sampling bias given that digital traces are being collected from mobile apps installed on smart phones, which limits the types of individuals from whom mobility data is being collected. In fact, prior work has revealed sampling bias in SafeGraph mobility data with Black and elder individuals being under-represented in the datasets  \cite{coston_bias_mob}.
Critical to the research proposed in this paper, prior work has exposed biases in COVID-19 forecasting models \cite{Tsai_Arik_Jacobson_Yoon_Yoder_Sava_Mitchell_Graham_Pfister_2022_covidfair},
and researchers have shown that COVID-19 county prediction improvements associated to the use of mobility data tend to take place 
in counties with lower presence of protected racial and ethnic groups \cite{abrar2023analysis}. On the other hand, COVID-19 under-reporting bias has been discussed in the literature \cite{douglas2021variation,gross2020racial} and points to multiple causes, including inadequate testing across certain minority groups such as Hispanic/Latinos \cite{ama}; or lack of consistency in reporting race and ethnicity for COVID-19 cases, which has generated a lot of missing or incorrect racial data in COVID-19 case statistics, as reported by the CDC \cite{cdcunderreporting}.
Reducing the impact of mobility or COVID-19 case bias in COVID-19 case predictions, as we do in this paper, is of critical importance to support decision making processes focused on resource allocation during pandemics, so as to reduce harms and guarantee that decisions are fair and just across racial and ethnic groups. 

\subsection{Fairness Metrics and Fairness Corrections in Machine learning models}

Transformer-based COVID-19 case forecast models require the use of fairness metrics for regression settings, given that the loss optimization process in gradient based deep learning architectures uses real number predictions instead of classes.
Agarwal et al. \cite{Agarwal_Dudík_Wu_2019_fairreg},  Fitzsimons et al. \cite{Fitzsimons_Al_Osborne_Roberts_2019_fairreg} or Gursoy et al. \cite{gursoy2022error} outline the different aspects of fairness in regression settings, and propose a set of fairness metrics for regression-type models. For this paper, we use the error parity metric proposed in \cite{gursoy2022error}. Error parity requires error distributions to be statistically independent from racial and ethnic groups. We expand this definition, and relax the statistical significance requirement, to be able to also evaluate whether the proposed DemOpts method can at least reduce the differences in error distributions across racial and ethnic groups, even when they are still be statistically significantly different.

To correct for bias and unfair performance in deep learning models, researchers have used pre-processing and in-processing correction approaches. Pre-processing approaches focus on creating a better input for learning deep neural network models by removing bias from the datasets \cite{GlOVEPreFair},\cite{OptoPre}; and there have been successful efforts focused on de-biasing under-reporting COVID-19 datasets to estimate actual cases or deaths before they are fed into predictive models \cite{jagodnik2020correcting, albani2021covid}.
On the other hand, in-processing approaches to improve the fairness of deep learning models, like the one we use in this paper, focus on the model and its regularization, usually adding a bias correction term in the loss function \cite{inproc_Socmedia, Das_Dooley_Inproc, Yan_Seto_Apostoloff_inproc}.

In this paper, we will compare our proposed de-biasing approach against three state-of-the-art methods for de-biasing in regression settings:  
Individual fairness correction~\cite{berk2017convex},
Group Fairness correction~\cite{berk2017convex} and sufficiency based penalty for fairness correction~\cite{sufficiencyabhin}.
Individual and group fairness calculate penalties by determining over-estimations across different groups and weighting the loss by a factor proportional to the over-estimations; while sufficiency based regularizers propose to make the loss independent of sensitive data attributes by simultaneously training a joint model and subgroup specific networks to achieve fair predictions~\cite{sufficiencyabhin}.

\section{DemOpts Method}

Our modeling focus is on deep learning models, which are the most frequently used approach for COVID-19 county case forecasts in the machine learning community \cite{meraihi2022machine}. We specifically focus on the Temporal Fusion Transformer (TFT) model introduced in \cite{TFT} for several reasons. First, this model is state of the art in interpretable time series prediction \cite{TFT}. 
Second, this model allows for the use of static reals as input to the model (attributes that do not change over the duration of the training process such as demographic percentages or population statistics); and third, the model works well with time-dependent features including COVID-19 cases or mobility data whereby past data influences future statistics. 

\begin{figure*}[h]
    \centering
     \includegraphics[width=\textwidth]{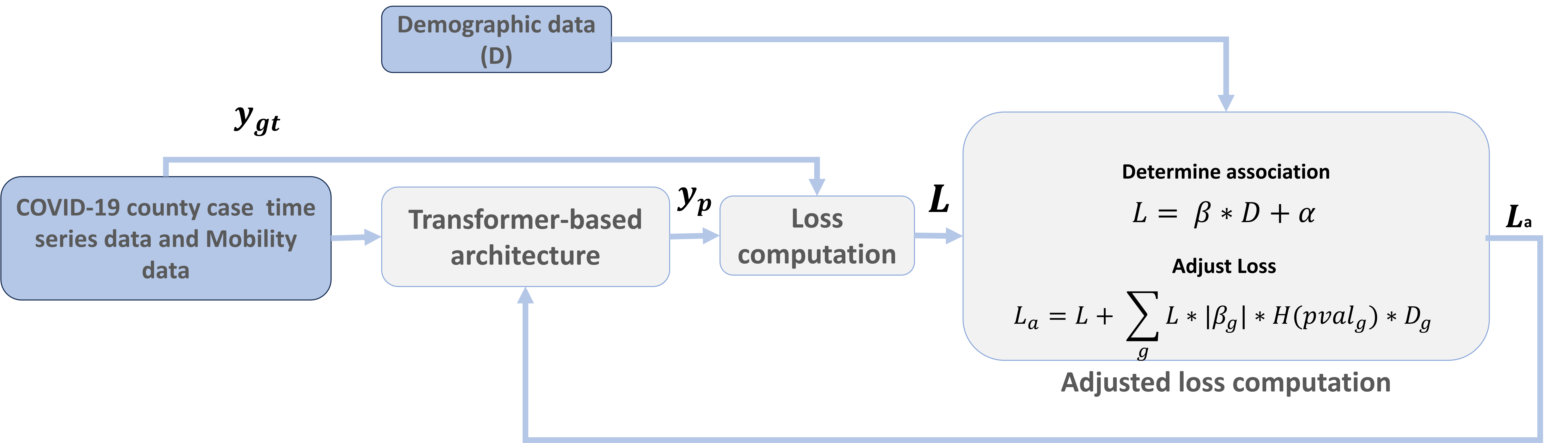}
    \caption{Flow Diagram for the DemOpts method. }
    \label{fig:teaserpic}
\end{figure*}

Training a deep learning model has the following steps: (1) forward pass on the training data, (2) computation of loss and (3) backward pass to change weights of the model. DemOpts modifies conventional loss functions to penalize the model for any statistically significant association between the county prediction loss (error) and the county racial and ethnic groups. In other words, DemOpts performs a race-based optimization on the error during model training using county demographic racial and ethnic data. To achieve that, DemOpts follows a three step process (see Figure \ref{fig:teaserpic} for a diagram and Algorithm \ref{algo:demopts} for the pseudocode):

\subsection{Step 1: Calculate Loss} 

To thoroughly model errors in time series, we use quantile predictions instead of point-value predictions.
Quantile predictions are measured for seven quantiles ([0.02, 0.1, 0.25, 0.5, 0.75, 0.9, 0.98]) to gain insights into the uncertainty ranges and confidence intervals of the COVID-19 county case predictive models. 
When using quantile predictions, the error is computed using the quantile loss, also known as pinball loss (PBL), and defined as follows:
\begin{equation}
    PBL_q(y_{ip}, y_i) = \begin{cases} q*(y_i - y_{ip}) & \text{if } y_i \geq y_{ip} \\
(q - 1)*(y_i - y_{ip}) & \text{if } y_i < y_{ip} \end{cases} 
\label{eq:pbl}
\end{equation}
For quantile $q$, the PBL for the prediction of a given input $X_i$ is $PBL_q(y_{ip}, y_i)$, where $y_i$ is the ground truth and $y_{ip}$ is the predicted value. The average over all quantiles can be represented as $PBL(y_{ip}, y_i) = \frac{1}{|q|}\sum_q PBL_q(y_{ip}, y_i)$.

\subsection{Step 2: Identify Dependencies between Prediction Errors and Race and Ethnicity} 
To achieve error parity \textit{i.e.,} mean errors being independent from racial and ethnic groups, DemOpts first determines 
the relationship between errors and race and ethnic labels.
For that purpose, DemOpts fits a regression model between the prediction losses $PBL(y_{ip},y_i)$ across datapoints $i$ and their corresponding county race and ethnicity labels $D_i$: 
\begin{equation}
\begin{alignedat}{2}
&PBL(y_{ip},y_i) &= \beta*D_i + \alpha \\
&& \qquad \text{with} \quad D_i &= \begin{bmatrix}d_1,d_2,d_3,d_4, \text{lookahead}\end{bmatrix}
\end{alignedat}
\label{step2TFTDemOpts}
\end{equation}

where $d_i$ are the corresponding county demographic features extracted from the U.S. census data and represented as the percentage of each racial and ethnic group 
of the county corresponding to datapoint $i$, 
and lookahead refers to the number of days into the future the COVID-19 case prediction was generated for.
In matrix representation:
$PBL(Y_{ip},Y_i) = \beta*D + \alpha$. Once the regression model is fit, both regression coefficients ($\beta$) and their statistical significance ($p-value$) are passed on to step 3, to modify the adjusted loss and attempt to decouple race from the errors (loss). 

\subsection{Step 3: Adjust the Loss} 
DemOpts modifies the conventional loss of deep learning models by adjusting for racial or ethnic bias in the error \textit{i.e.,} the loss is increased whenever a statistically significant regression coefficient for a race or ethnicity is found in Step 2 (at level $p-value = 0.05$).
By increasing the loss, DemOpts attempts to reduce the dependency between errors and race and ethnicity \textit{i.e.,} make the errors similar across racial and ethnic groups.
Specifically, the loss is adjusted by the product of the prior loss $PBL(y_{ip},y_i)$, the percentage race or ethnicity $D_j$ that holds a significant relationship with the error, and its coefficient $\beta_j$ in absolute value: 
\begin{equation}
\begin{alignedat}{2}
L_{adj} = PBL(y_{ip},y_i) + \sum_j H(pval_j)(|\beta_j|\*D_j\*L) \\
\text{with} \qquad H(x) = \begin{cases}
1 & \text{if } x < 0.05, \\
0 & \text{if } x \geq 0.05
\end{cases}
\end{alignedat}
\label{tag:step3TFTDemOpts}
\end{equation}

\begin{algorithm}[ht]
\begin{myframe}
\caption{DemOpts Training}
\label{algo:demopts}
\begin{algorithmic}[1]
\scriptsize
\State \textbf{Input:} Training set (X, D, Y), Learning rate (lr), Number of epochs (epochs), threshold
\State \textbf{Output:} Trained model (M)
\State X : COVID-19 Timeseries data for all counties
\State Y : COVID-19 cases in future for all counties
\State D : Demographic data for all counties  
\State Initialize model parameters randomly
\For{epoch in range(0, epochs)}
    \State // sample from $X,D,Y$ of size b
    \For{$(X_b,D_b, Y_{bt})$ in $(X, D, Y)$}
        \State // Forward propagation
        \State $Y_{bp} = M(X_b)$
        \State //Calculate QuantileLoss
        \State $L_b = QuantileLoss(Y_{bp}, Y_{bt})$
        \State //Find association
        \State $olsreg = OLS.fit(D_b,L_b)$
        \State $pvals,betas = olsreg.pvals,olsreg.coef$
        \State // additional penalty on loss
        \For{$index$ in $|pvals|$}
            \State $pval_i, beta_i = pvals[index], betas[index]$
            \State // Get the corresponding demographic percentage column and all rows 
            \State $D_{b,idx} = D_b[:,index]$
            \If{$pval_i < threshold$} //this ensures significant association
                \State $L_b += L_b*|beta_i|*D_{b,idx}$  
            \EndIf
        \EndFor
        \State $backpropagate(M,L_b)$   
    \EndFor
\EndFor
\State \textbf{return} TFT
\end{algorithmic}
\end{myframe}
\end{algorithm}

\section{DemOpts Evaluation Protocol}
In this section, we describe a novel protocol to evaluate DemOpts. For that purpose, we first describe the COVID-19 county case prediction model we use, and the different de-biasing approaches we evaluate on that prediction model. Next, we describe the error parity metrics we use to evaluate the fairness of each prediction model; and finally, we present the approach to analyze whether DemOpts improves the error parity metrics when compared to other state-of-the-art de-biasing approaches for regression settings. 

 \subsection{Predictive Model and De-Biasing Approaches}
We use the Temporal Fusion Transformer model (TFT) with the conventional pinball loss function as our baseline model ($TFT_{baseline}$)
to predict the number of COVID-19 county cases for a given day. 
Input data to the TFT model include past COVID-19 cases per county, mobility data from SafeGraph and race and ethnicity data for the county (further details about these datasets are provided in the next section).

We also train and test another TFT enhanced with 
the DemOpts de-biasing method, $TFT_{DemOpts}$, that adjusts the loss computation to attempt to eliminate or reduce the dependencies between error and race so as to achieve error parity. 
In addition, we train and test three more TFTs enhanced with state-of-the-art de-biasing methods for  regression settings namely, individual fairness $TFT_{Individual}$~\cite{berk2017convex},
group fairness $TFT_{Group}$~\cite{berk2017convex},
and 
sufficiency based regularizer $TFT_{Sufficiency}$~\cite{sufficiencyabhin}.
Individual and group fairness methods calculate penalties by determining over-estimations across different groups and weighting the loss by a factor proportional to the over-estimations; while the sufficiency based regularizer
trains a joint model and group-specific networks to achieve fair predictions. We replicate their methodology and adapt it to the forecasting setting by keeping TFT as the common network in the training process.

\subsection{Measuring Model Fairness}
We choose error parity as our fairness metric \cite{gursoy2022error} with a focus on evaluating whether the distribution of predictive errors at the county level is independent of county race and ethnicity \textit{i.e.,} prediction errors are not statistically significantly different across racial and ethnic groups. To measure the fairness of each of the models $TFT_{baseline}$, $TFT_{DemOpts}$, $TFT_{Individual}$, $TFT_{Group}$ and $TFT_{Sufficiency}$, we propose a two-step process. 

\begin{table}
    \centering
    \caption{Majority label counts }
    \begin{tabular}{|c|c|}
         \hline
         Majority label&  Count\\
         \hline
         Asian    &6\\
         Black    &127\\
         Hispanic &126\\
         White &2825\\
        \hline
    \end{tabular}
    \label{tab:majtable}
\end{table}

\begin{figure*}
    \centering
     \includegraphics[width=0.75\textwidth]{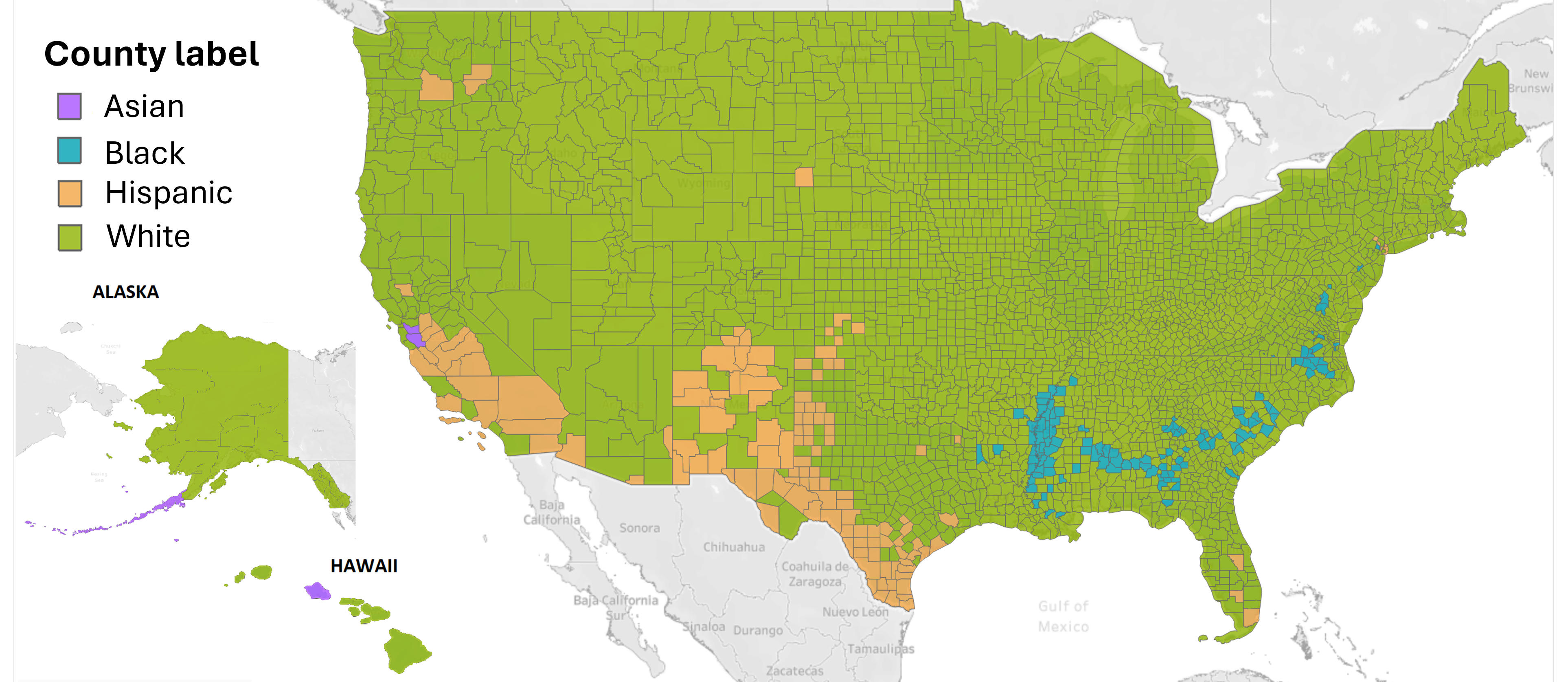}
    \caption{Counties with majority based labels.}
    \label{fig:plurality}
\end{figure*}

\textbf{Step 1: Associate errors to county race or ethnicity.} To carry out the fairness analysis, we need to associate the PBL error of each county with race and ethnicity labels. 
However, that would require access to race-stratified COVID-19 case data at the county level, unfortunately not available due to systemic data collection failures during the pandemic~\cite{kader2021participatory}. Hence, we propose to associate each county and its error to the majority race: we label each county with the race or ethnicity that has the highest population percentage in that county. 
Following this procedure and using data from the 2019 U.S. Census, our fairness analysis will consider the following race and ethnic groups: Asian, Black, Hispanic and White. Table \ref{tab:majtable} shows the distribution of U.S. counties into these four racial and ethnic groups, and Figure \ref{fig:plurality} shows a color-coded map with the majority racial or ethnic group for each county. During the fairness analysis, we will refer to majority White counties as the unprotected group, and majority Black, Hispanic or Asian counties as the protected groups.

In addition, we normalize 
each county PBL error by 1,000 county population people. The normalization by county population allows us to scale the errors appropriately, since higher population counties will have higher case counts and thus, higher magnitude of error. Normalizing by population fairly compares error per unit population of one county with another. 

\begin{equation}    \label{eq:normpbl}
\begin{aligned}
NormPBL(y_{pi},y_{ti}) &= \frac{1000*PBL(y_{pi},y_{ti})}{pop_i}\\
\end{aligned}
\end{equation}
where $y_i$ is the ground truth, $y_{ip}$ is the predicted value, and $pop_i$ is the county population.

\textbf{Step 2: Compute fairness metric.} Once PBLs have been associated with racial and ethnic groups in the U.S., we can compute the error parity \textit{i.e.,} the fairness metric focused on evaluating whether the prediction errors are different across race and ethnicity. 
We propose two metrics to measure the error parity of COVID-19 county case predictions: hard error parity and soft error parity. Next, we explain how we implement them and why both are needed. 

\textit{Hard Error Parity Metric.} Model predictions exhibit hard error parity when no statistical significant difference exists between county case normalized mean prediction errors (NormPBL) across racial or ethnic groups. In other words, 
normalized mean PBL errors across counties of 
different racial and ethnic groups are similar and hence not biased by race or ethnicity. To test for the hard error parity of a prediction model, we propose to run one-way ANOVA followed by post-hoc Tukey HSD tests between the county normalized mean error distributions of all racial and ethnic groups.
ANOVA tests have been shown to be an adequate choice even in violation of normality, and in the presence of unequal sample sizes, 
like our majority race/ethnic distributions;
thus, we choose this parametric test due to its superior strength \cite{blanca2017non, zimmerman1987comparative}.

Rejecting the null hypothesis for ANOVA would point to significantly different mean error values across some racial or ethnic groups, and to a lack of \textit{perfect hard error parity}.
The subsequent analysis of the post-hoc Tukey-HSD test would reveal the pairs of racial and ethnic groups whose mean error values are significantly different, and their numerical difference. The Tukey test also highlights the pairs of racial and ethnic groups for which the mean error is not statistically significantly different, pointing to instances where hard error parity exists for that model.
However, as Table \ref{tab:majtable} shows, the number of points in some distributions might not be sufficient to reveal statistically significant results \cite{samplesize}. 
Thus, we also propose a relaxed definition of error parity: soft error parity.

\textit{Soft Error Parity Metric.} Instead of measuring the statistical significance of the relationship between county race labels and county errors, we propose to use the Accuracy Equity Ratio metric (AER) \cite{castelnovo2022clarification}. AER computes the ratio between the errors of the protected and unprotected groups as follows:

\begin{equation} 
\begin{aligned}
AER_{pg} &= \frac{ AvgNormPBL(y_{p},y_{t},pg)}{AvgNormPBL(y_{p},y_{t},unpg)} 
\end{aligned}
\label{eq:AER}
\end{equation}

where subscript $pg$ indicates counties labeled as protected group (Black, Hispanic or Asian), $unpg$ indicates counties labeled as the unprotected group (White), and
AvgNormPBL is the average of the normalized PBL across all counties for a given racial group $g$ ($pg$ or $unpg$):

\begin{equation}
\begin{aligned}
AvgNormPBL(y_{p},y_{t},g) &= \sum_{i \in c_g} \frac{ NormPBL(y_{pi},y_{ti})}{\lvert c_g \rvert}
\end{aligned}
\end{equation}

As defined, the $AER$ metric goes from $0$ to $\infty$. $AER$ values in the range $0$ to $1$ indicate comparatively lower normalized PBL  for protected groups, which means the model predictions could be biased - have higher errors - for White majority counties; while $AER$ values larger than one indicate that the model could be biased against the protected group \textit{i.e.,} the prediction errors are larger for counties with majority Black, Hispanic or Asian population. Values close to one indicate parity in error distribution between the protected group counties and the majority White counties. We claim that a predictive model achieves soft error parity for a given protected group when the $AER$ value is close to one, that is, the mean predictive error between that protected group and the White race is similar.

\subsection{DemOpts Over State-of-the-Art}
To assess whether DemOpts is a better de-biasing approach than state-of-the-art methods, we need to compare the error parity metrics of the COVID-19 county case prediction model enhanced with the DemOpts method $TFT_{DemOpts}$ against the error parity metrics of the same prediction model enhanced with the other de-biasing approaches (individual $TFT_{Individual}$, group $TFT_{Group}$ or sufficiency $TFT_{Sufficiency}$) as well as with the baseline COVID-19 county case prediction model without any de-biasing approach, $TFT_{Baseline}$. 

Next we describe how we carry out this analysis for the hard and soft error parity metrics.

\textit{Hard Error Parity Evaluation.} 

We compute the hard error parity metric for each of the COVID-19 county case prediction model, using one-way ANOVA and the post-hoc Tukey-HSD test. An exploration of the statistical significance of the mean error difference for each pair of racial and ethnic groups will reveal whether 
applying DemOpts to the COVID-19 case prediction model produces less instances of significant 
mean prediction error differences than any of the other de-biasing methods or the baseline. 
In other words, a decrease in the number of significantly different mean PBL errors between races would point to an achievement of
hard error parity for more racial and ethnic groups that other state-of-the-art de-biasing approaches, or than the baseline.

\textit{Soft Error Parity Evaluation.} To assess whether DemOpts applied to a COVID-19 case prediction model has higher soft error parity
than any of the other state-of-the-art de-biasing approaches, we propose to compare the AER values for each protected race and ethnic group across the five models: $TFT_{DemOpts}$, $TFT_{Individual}$, $TFT_{Group}$, $TFT_{Sufficiency}$ and $TFT_{Baseline}$.
Since AER values represent the quotient between the normalized mean prediction errors of a protected race/ethnicity versus White counties,
the model with more AER values closer to one will be the approach with the highest soft error parity. 
To measure AER's \textit{distance to one}, we compute the $distance = |1-AER_{race}|$ for each race and ethnic group, which represents the distance to a perfect soft parity error of $1$. Distances closer to zero reveal better soft parities \textit{i.e.,} soft parity values closer to one.

\section{DemOpts Evaluation Results}
We first present the datasets and the models used to train and test the COVID-19 county case prediction models with and without de-biasing approaches. We finalize discussing the hard and soft error parity analysis for $TFT_{DemOpts}$ by comparing it against the other state-of-the-art de-biasing methods and against the baseline. 

\subsection{Datasets}

In this paper, we train TFT COVID-19 county case prediction models for the U.S. using 
COVID-19 case data, as well as mobility and demographic data. Mobility data has been used by prior work in an attempt to inform case prediction via human mobility behaviors, under the assumption that the way people move might have an impact on the spreading of the epidemic. On the other hand, demographic data either raw from the census, or combined in different types of vulnerability indices, has also been shown to be helpful in predicting COVID-19 prevalence, given the fact that COVID-19 
has heavily affected vulnerable populations \cite{gross2020racial}.

\textbf{COVID-19 Case Data.}
We use the COVID-19 case data compiled by the New York Times at the county level~\cite{nytimes2021covid}. 
We account for delayed reporting, by using the 7-day daily rolling average of COVID-19 cases (computed as the average of its current value and 6 prior days) instead of raw counts. 
As stated in the Introduction, case numbers might not be reflective of the actual spread of COVID-19 for specific racial and ethnic groups, and such under-reporting bias could in turn affect the fairness of the COVID-19 predictions \cite{douglas2021variation, ama}.

\begin{figure}[ht]
\centering
\begin{subfigure}{0.5\textwidth}
    \includegraphics[width=\textwidth]{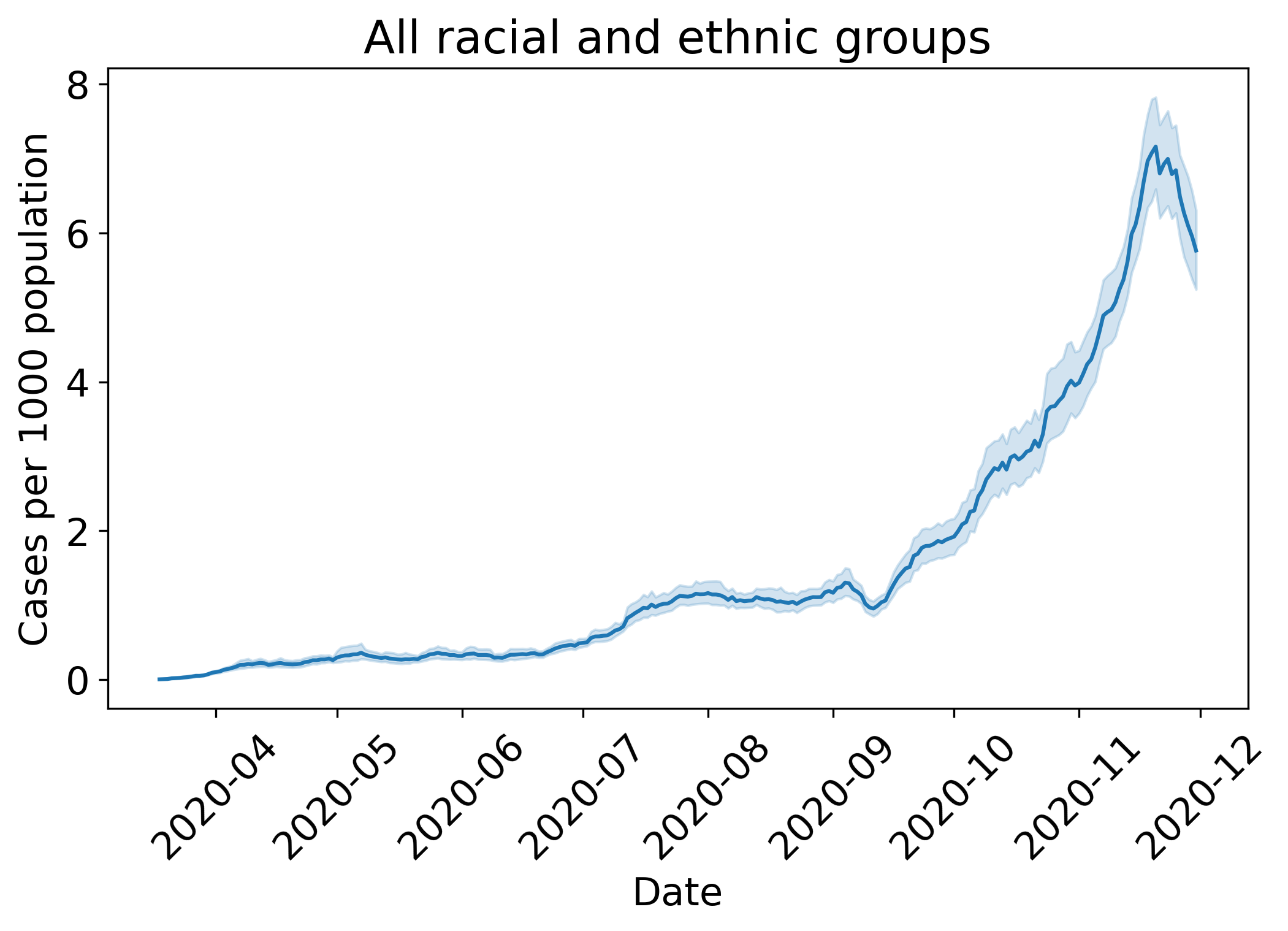}
    \caption{}
    \label{fig:case_per1000}
\end{subfigure}

\vspace{0.5em}  

\begin{subfigure}{0.5\textwidth}
     \includegraphics[width=\textwidth]{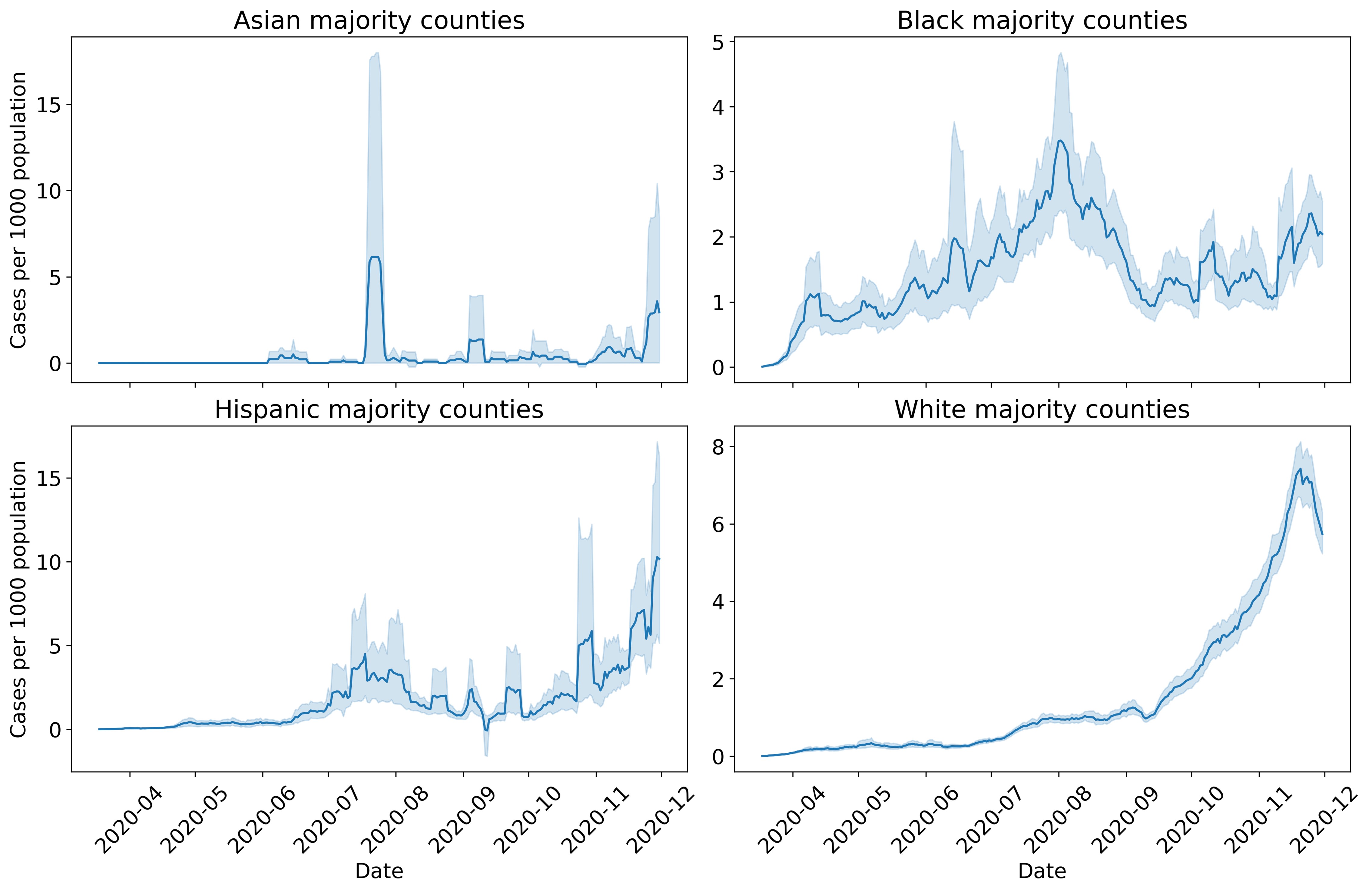}
    \caption{}
    \label{fig:per_race_case}
\end{subfigure}
\caption{(a) Case counts per 1000 population for all ethnicities and race. (b) Case counts per 1000 population using majority based labelling of counties. \textit{Note scale difference in y-axis.}}
\end{figure}

\begin{figure}
    \centering
     \includegraphics[width=0.5\textwidth]{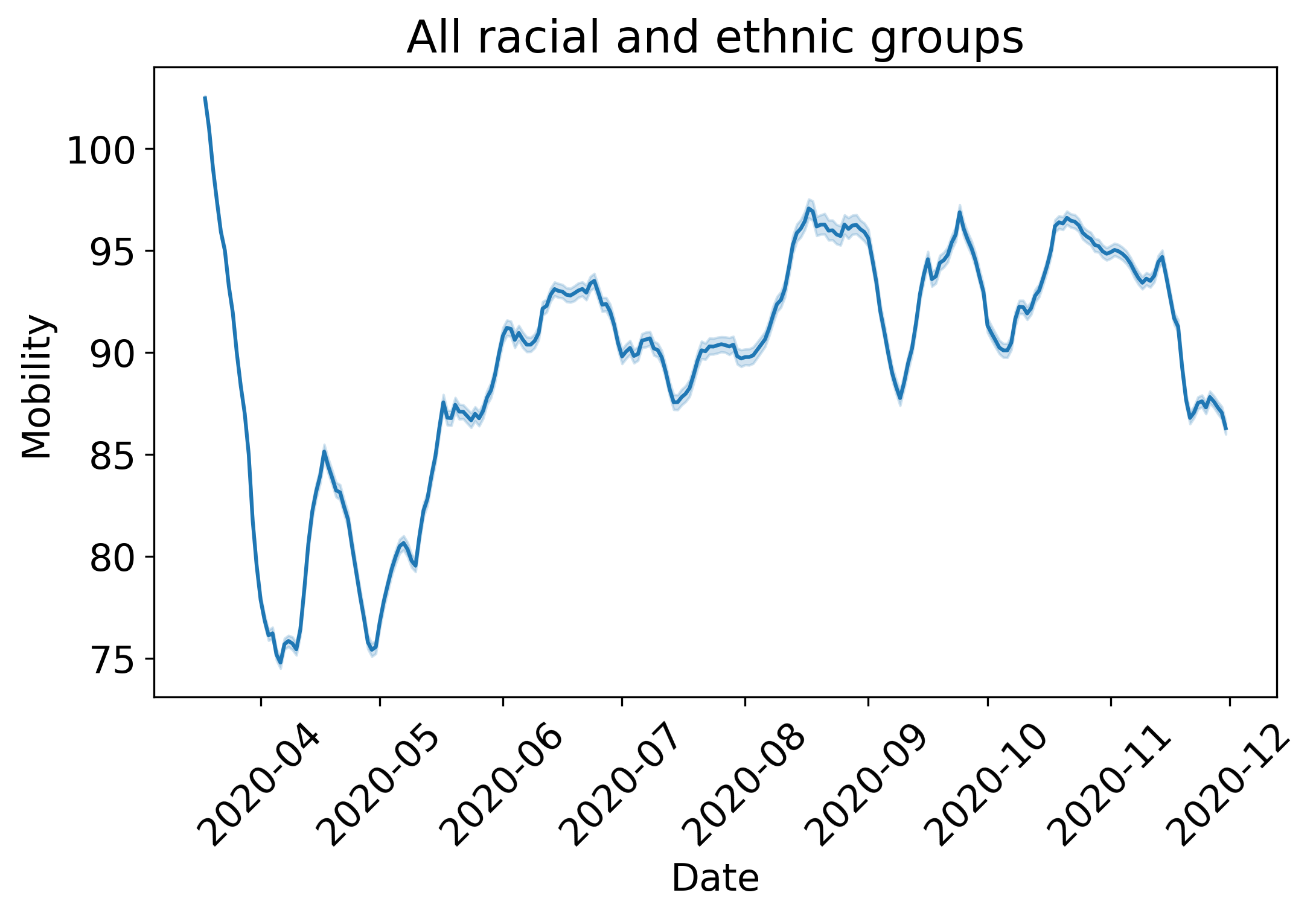}
    \caption{Mobility for all ethnicity and races.}
    \label{fig:mobility_all}
\end{figure}

Figure~\ref{fig:case_per1000} reflects the daily COVID-19 reported cases throughout the data collection period and Figure~\ref{fig:per_race_case} shows the case temporal distribution per race and ethnicity.

\begin{table}[!htbp]
\centering
\begin{tabular}{ll}
\toprule
       Fairness Method & F(x,y)  \\
\midrule
 Baseline &  1195.398** \\
          Demopts &   668.769**  \\
     Group &  1455.528**  \\
     Individual &  1469.698**  \\
          Sufficiency &  1195.651**  \\
\bottomrule
\end{tabular}
\caption{ ANOVA F-test statistic comparing the mean prediction error for each TFT prediction model: baseline and TFTs enhanced with de-biasing methods. All tests were significant with p-value $< 0.01$ (**) with DF (3,3080).}
\label{tab:anovaftab}
\end{table}

\textbf{SafeGraph Mobility Data.}
SafeGraph open sourced the mobility patterns of smart phone app users at the onset of the pandemic. These data points are curated by tracking the movements of millions of pseudonymized users via mobile app SDKs. Based on the data available, we use the daily O-D (origin-destination) county to county flows \cite{safegraphdata}. O-D flows represent the volume of trips between pairs of counties across the United States for each day. For O-D flows, we only use SafeGraph inflow (i.e. the mobility to the county). The inflow mobility is measured as changes in volumes of flows with respect to a baseline of normal behavior computed by SafeGraph using mobility data between the period February 17, 2020 to March 7, 2020.
Prior work has shown sampling bias in mobility datasets revealing that not all races and ethnicities are equally represented \cite{coston_bias_mob,Schlosser_bias_mob}. It has also been shown that sampling bias in the mobility data can negatively impact downstream tasks such as COVID-19 forecasting \cite{abrar2023analysis}. While the addition of mobility could potentially help improve the prediction accuracy and support better decision making, it introduces bias. Our empirical analysis of DemOpts aims to understand whether the de-biasing method proposed in this paper can improve the fairness of COVID-19 county case predictive models when mobility data is used as input into the predictive model. Figure \ref{fig:case_per1000} depicts the daily average mobility across all counties in the US throughout the data collection period.

\textbf{Race and Ethnicity Data.}
We retrieve the race and ethnicity data from each county in the U.S. from the American Community survey (ACS). The ACS survey collects data annually from all 50 states, Puerto Rico and Washington DC. We use the population race and ethnicity information for each county, and consider the following labels: Asian, Black, Hispanic and White. 

\begin{table*}[!htbp]
\centering
\begin{tabular}{ll|lllll}
\toprule
         & Fairness Method &  Baseline &   DemOpts &     Group & Individual & Sufficiency \\
Group1 & Group2 &           &           &           &            &             \\
\midrule
Asian & Black &    -0.118 &     1.327 &    -0.202 &     -0.126 &      -0.119 \\
         & Hispanic &  -2.302** &   \textbf{ -0.771} &  -2.659** &   -2.507** &    -2.297** \\
         & White &  -2.064** &    \textbf{-0.968} &  -2.515** &   -2.517** &    -2.061** \\
Black & Hispanic &  -2.184** &  -2.098** &  -2.457** &   -2.381** &    -2.178** \\
         & White &  -1.946** &  -2.295** &  -2.313** &   -2.391** &    -1.942** \\
Hispanic & White &     0.238 &    -0.197 &     0.144 &      -0.01 &       0.236 \\
\bottomrule
\end{tabular}

\caption{Hard error parity analysis. Each number represents the difference between the mean normalized PBL loss for each pair of racial and ethnic groups, and its statistical significance \textit{i.e.,} whether the difference is statistically significant or not (with ** p-value $< 0.01$, * p-value $< 0.1$). Bolded numbers represent the instances where DemOpts has removed a significant difference in mean PBL errors, improving over the baseline and all the other de-biasing methods. DemOpts achieves hard error parity for Asian counties. }
\label{tab:hardparityoveralldiff}
\end{table*}
\subsection{COVID-19 Case Prediction Models}
We use COVID-19 case data as well as SafeGraph mobility data from March 18, 2020 to November 30, 2020 for the training (207 days) and testing (49 days) of the TFT COVID-19 county case prediction models. 
The forecast task is the prediction of the number of COVID-19 cases for a given county for day $X + 1$ to $X + 49$ \textit{i.e.,} the following 3 months (long-term forecasting with lookahead values from $1$ to $49$).
Specifically, we train and test: (1) the $TFT_{baseline}$, a TFT prediction model without a de-biasing method; (2) the $TFT_{Individual}$, $TFT_{Group}$ and $TFT_{Sufficiency}$, TFT prediction models with state-of-the-art de-biasing methods and (3) $TFT_{DemOpts}$, a TFT prediction model enhanced with our proposed de-biasing method.  
All five models are trained and tested for exactly the same temporal range; and all are implemented using the pytorch-forecasting library.
Although COVID-19 county case data as well as mobility data are available for longer periods of time, we decided to limit the period of
analysis to a time before COVID-19 vaccines were available, given that after that event, research has revealed a very unclear relationship between mobility data and post-vaccines COVID-19 case volumes \cite{gatalo2021associations_covidmobrelation}.
Once all models have been trained, we use the prediction errors (PBL) per racial and ethnic group to analyze and compare their hard and soft error parity.

\begin{table*}[!htbp]
\centering
\begin{tabular}{l|rrrrr}
\toprule
Fairness Method & Baseline & Demopts &  Group & Individual & Sufficiency \\
Group    &          &         &        &            &             \\
\midrule
Asian    &    0.811 &   \textbf{0.248} &  0.842 &      0.850 &       0.811 \\
Black    &    0.764 &   \textbf{0.588} &  0.774 &      0.807 &       0.764 \\
Hispanic &    0.093 &   0.051 &  0.048 &      \textbf{0.003} &       0.093 \\
\bottomrule
\end{tabular}
\caption{Soft error parity analysis. Each number represents the distance ($|1-AER_{race}|$) for each protected group and de-biasing method. For each protected race/ethnicity, distances closer to zero represent higher soft error parity (signaled in bold font). $TFT_{DemOpts}$ achieves the highest soft error parity for two of the three protected races under study. $TFT_{Individual}$ achieves the best soft error parity for the Hispanic counties when compared to prediction errors in White counties.    }
\label{tab:unfairnesstab}

\end{table*}

\subsection{Hard Error Parity Analysis} 
ANOVA tests of the normalized mean PBL error distributions across race and ethnic groups for each de-biasing approach were all significant, pointing to a dependency between race and the normalized prediction errors. 
Table \ref{tab:anovaftab} shows the F-statistic and test significance for each of the prediction models with and without de-biasing approaches.
The significant ANOVA tests reveal that perfect hard error parity is not achieved by any of the de-biasing methods. 
In other words, for some racial and ethnic groups there exist statistically significant differences between their mean PBL prediction errors and those of other racial and ethnic groups; and this effect happens for the $TFT_{baseline}$ as well as across all the other predictive models enhanced with a de-biasing approach. 

Nevertheless, 
post-hoc Tukey-HSD tests revealed interesting nuanced results, showing significant differences in errors only between specific pairs of racial and ethnic groups. 

Table~\ref{tab:hardparityoveralldiff} shows the post-hoc Tukey-HSD test results for each COVID-19 case predictive model: the baseline and each of the four models enhanced with a de-biasing approach. 
Each row represents the output of the post-hoc test \textit{i.e.,} the difference between the normalized mean PBL error of Group 1 and Group2 \textit{i.e.,} $NormPBL_{Group1}-NormPBL_{Group2}$. If the difference is positive, it means that the normalized mean predictive error is higher for Group 1; if the difference is negative, the normalized PBL is higher for Group 2. The asterisks indicate whether the difference is statistically significant or not.

The first relevant observation in looking at the table is that the baseline model, focused on predicting COVID-19 county cases with no de-biasing approach, is highly biased, with statistically significant differences between the mean normalized errors across all pairs of races, except for the comparison between Asian and Black counties as well as Hispanic and White counties, for which there is no statistically significant difference between the prediction errors. 
These results reveal that there is no racial group or ethnicity that achieves hard error parity, and motivates
 our exploration of whether state-of-the-art de-biasing methods or our proposed DemOpts can improve the hard error parity results of the baseline model.

Looking at Table~\ref{tab:hardparityoveralldiff}, we can observe that predictive models enhanced with the Individual, Group, or Sufficiency de-biasing methods do not improve the hard error parity over the baseline. In fact, each pair of racial and ethnic groups whose prediction error distributions are significantly different 
for the baseline (rows with asterisks in the Baseline column), remain significantly different for the Individual, Group and Sufficiency de-biasing methods (rows with asterisks in the Individual, Group and Sufficiency columns).
Looking at the significant mean PBL differences between racial and ethnic groups for the baseline and the state-of-the art de-biasing models, we observe that all coefficients have similar values, signaling similar significant mean PBL differences between racial and ethnic groups (with values between $1.942$ and $2.659$ error cases by 1,000 population). The sign of the coefficients reveals higher mean PBL errors for Hispanic and White counties when compared to Asian counties or Black counties; and higher mean PBL errors for White counties when compared to Hispanic counties across all five models (baseline and four de-biasing approaches). For example, Hispanic and White counties have mean prediction errors $2.302$ and $2.064$ higher, when compared to Asian counties and while using the baseline model; and Hispanic and White counties have errors $2.184$ and $1.946$ higher when compared to Black counties and while using the baseline model. 
\begin{table*}
\centering
\begin{tabular}{l|rrrrr}
\toprule
Fairness Method & Baseline & Demopts &  Group & Individual & Sufficiency \\
Group    &          &         &        &            &            \\
\midrule
Asian    &    0.482 &   2.938 &  0.472 &      0.444 &       0.479  \\
Black    &    0.600 &   1.611 &  0.674 &      0.570 &       0.598 \\
Hispanic &    2.784 &   3.709 &  3.131 &      2.951 &       2.776 \\
White    &    2.546 &   3.906 &  2.987 &      2.961 &       2.540  \\

\bottomrule
\end{tabular}

\caption{Average prediction error (PBL) for each racial and ethnic group and for each TFT COVID-19 county case prediction model: baseline model, and models enhanced with a de-biasing method (Individual, Group, Sufficiency and DemOpts). DemOpts achieves fairness by increasing mean errors for the Asian and Black groups.}
\label{tab:pinball_smape}
\end{table*}
Notably, all predictive models including the baseline and those enhanced with a de-biasing method ($TFT_{DemOpts}$, $TFT_{Group}$, $TFT_{Individual}$ and $TFT_{Sufficiency}$) achieve hard error parity between Asian and Black counties and between Hispanic and White counties {\it i.e.,} the mean error difference between these counties is not significant. 
But even more interesting is the fact that DemOpts is the only de-biasing method that achieves hard error parity in more cases than the baseline, effectively removing some of the associations between race and ethnicity and the normalized mean error distribution (PBL). Specifically, DemOpts removes the significant difference between the prediction errors of Asian and White counties, and of Asian and Hispanic counties (see bolded values in the Table), effectively achieving hard error parity for Asian counties {\textit i.e.,} the mean PBL in Asian counties is always similar to the mean error in counties of all the other racial and ethnic groups. 
And these DemOpts improvements take place
while maintaining the $TFT_{baseline}$ hard error parity between Asian and Black and Hispanic and White counties - also present in the other three de-biasing methods. In other words, \textbf{DemOpts improves the hard error parity of COVID-19 county case predictions for two more racial and ethnic pairs than any of the 
other de-biasing methods.}

Finally, when looking specifically into the hard error parity between protected (Asian, Black and Hispanic) and unprotected groups (White), DemOpts achieves hard error parity for Asian and Hispanic \textit{i.e.,} their mean prediction errors are not significantly different with respect to the White race; while the baseline and the other three de-biasing methods only achieve hard error parity for the Hispanic group when compared to White. These findings with respect to the White group lead us to evaluate the soft error parity of the different models, to understand, for example, if DemOpts achieves the best soft error parity for the Black group (since hard error parity was not achieved), or to see if DemOpts has better soft error parity than other de-biasing methods for Asian or Hispanic groups. Next, we explore the soft error parity metric for the TFT baseline and for all TFT models enhanced with de-biasing approaches. 

\subsection{Soft Error Parity Analysis} 
Table \ref{tab:unfairnesstab} shows the distance to the perfect soft error parity for each of the de-biasing approaches and across all protected racial and ethnic groups. 
As we can observe, DemOpts has the smallest values - closest distances to perfect soft error parity - for Asian and Black counties; while the Individual de-biasing method almost achieves perfect soft error parity for the Hispanic counties. In other words, the errors for Asian and Black counties are the closest to errors in White counties for the proposed DemOpts method, while the Individual de-biasing model achieves errors for Hispanic counties that are the closest to the White group.
In addition, it is important to highlight that the Group and Sufficiency de-biasing methods achieve soft error parities that are similar to the $TFT_{baseline}$ which is not enhanced with any de-biasing method.
Overall, \textbf{these results reveal that DemOpts is the de-biasing approach that improves the most the soft error parity of COVID-19 county case prediction models, with errors for Asian and Black counties being the closest to errors in White counties; while the Individual de-biasing method achieves the closest errors to the White race for Hispanic counties only.} 

\subsection{Why is DemOpts better?}
The results have shown that DemOpts is the only de-biasing approach to achieve both hard or soft error parity for all three racial minority groups when compared to the White race.
In an attempt to understand why DemOpts succeeds in increasing both hard and soft error parity in the context of COVID-19 county case predictions, and when compared to other de-biasing methods, we computed the average PBL for each racial and ethnic group and for each predictive model enhanced, or not, with a de-biasing method (see Table~\ref{tab:pinball_smape}). 

We can observe that DemOpts achieves better hard and soft error parity metrics because it considerably increases the errors for Asian and Black counties with respect to the baseline, until the differences with Hispanic and White are made not statistically significant (hard error parity) or closer to the White mean errors (soft error parity). This result also points to another interesting insight: the fact that DemOpts' optimization could not decrease prediction errors while trying to improve fairness, when fairness is measured via statistical significance,  showing a fairness-accuracy trade-off that has been reported previously in the literature~\cite{kim2020fact}. 
Finally, it is also important to clarify that, in practice, the prediction error increases brought about by DemOpts are
not that large, with increases between $1-2.5$ error cases by 1,000 people.
We posit that these small increases are acceptable if that is the requirement to guarantee hard and soft error parity across protected and unprotected racial and ethnic groups.

\section{Conclusion}
In the past four years, researchers have worked profusely on the creation of accurate COVID-19 case prediction models using not only historical COVID-19 cases but also complementary data such as human mobility or socio-demographic information. 
However, there exists prior work showing that the accuracy of COVID-19 predictions can depend on various social determinants,
including race and ethnicity, income, or age, revealing worse performance for protected attributes and pointing to
a lack on COVID-19 predictive fairness that could affect resource allocation and decision making.

In this paper, 
we show that state of the art architectures in COVID-19 case predictions (TFT models) incur in unfair prediction error distributions, and we design a novel de-biasing approach to increase the fairness of the predictions in the context of COVID-19 county case predictions.
The new proposed de-biasing approach, DemOpts, modifies the loss function in deep learning models to reduce the dependencies between error distributions and racial and ethnic labels. Our results show that
DemOpts improves the most both the hard and soft error parity of COVID-19 county case predictions when compared to state-of-the-art de-biasing methods.

\bibliography{aaai24}

\end{document}